\documentclass[11pt]{article}

\usepackage[final]{acl}

\usepackage{times}
\usepackage{latexsym}
\usepackage[T1]{fontenc}
\usepackage[utf8]{inputenc}
\usepackage{microtype}
\usepackage{inconsolata}

\usepackage{amsmath,amssymb}
\usepackage{bbm}
\usepackage{booktabs,array,tabularx,makecell,multirow}
\usepackage{graphicx}
\usepackage{enumitem}
\usepackage{algorithm}
\usepackage[noend]{algpseudocode}
\usepackage{balance}
\usepackage[table]{xcolor}
\usepackage{colortbl}
\usepackage{float}
\usepackage[most]{tcolorbox}
\usepackage{listings}
\lstdefinestyle{prompt}{
  basicstyle=\scriptsize\ttfamily, breaklines=true, columns=fullflexible,
  keepspaces=true, frame=single, framesep=4pt, rulecolor=\color{black!30},
  backgroundcolor=\color{black!3}, aboveskip=4pt, belowskip=4pt,
  xleftmargin=2pt, xrightmargin=2pt
}

\newcolumntype{Y}{>{\raggedleft\arraybackslash}X}
\newcolumntype{L}{>{\raggedright\arraybackslash}X}
\setlist[itemize]{leftmargin=*,topsep=2pt,itemsep=1pt}
\setlist[enumerate]{leftmargin=*,topsep=2pt,itemsep=1pt}

\newcommand{\method}{\textsc{GraphSU}}
\newcommand{\llamamodel}{\textsc{Llama-3.2-3B-Instruct}}
\newcommand{\llamacheckpoint}{\texttt{meta-llama/Llama-3.2-3B-Instruct}}
\newcommand{\softleak}{\textsc{SoftLeak}}
\newcommand{\listsummary}{\mbox{list-sum.}}
\newcommand{\relinversion}{\mbox{relation-inversion}}
\providecommand{\Description}[1]{}
\newcommand{\figmaybe}[2][]{\IfFileExists{#2}{\includegraphics[#1]{#2}}{\fbox{\begin{minipage}[c][0.16\textheight][c]{0.94\linewidth}\centering Missing figure file:\\ \texttt{\detokenize{#2}}\end{minipage}}}}

\title{Graph-Guided Selective Unlearning for Language Models:\\Controlling Support Routes Beyond Forget Seeds}

\author{Waqas Khan, Tabinda Sarwar, Jingyue Cong, Xun Yi, and Estrid He}

\begin{document}
\maketitle

\begin{abstract}

Enterprises fine-tune language models on proprietary data that may later require removal due to privacy, contractual, or compliance obligations. Selective unlearning removes requested knowledge while preserving model utility, offering a practical alternative to full retraining, but existing methods treat the explicitly identified forget examples as the complete deletion scope. This is insufficient when target knowledge remains recoverable through paraphrases, aliases, or neighboring training examples.
We propose \method, a graph-guided controller that expands the deletion scope beyond forget seeds by constructing a weighted support-route graph, propagating deletion pressure through it, and applying graded forgetting strengths to high-risk neighbors. On the Task of Fictitious Unlearning (\textsc{TOFU}), a synthetic author-profile question--answering benchmark, and \textsc{PISTOL}, a structural-unlearning benchmark built around interconnected factual samples, with GPT-2 Medium and Llama-3.2-3B-Instruct, \method{} achieves the lowest utility-feasible soft leakage across all deletion settings, reducing leakage by up to 49.5 percentage points over a matched seed-only baseline, demonstrating that effective enterprise unlearning requires controlling support routes, not just forget seeds. Our code is available at \url{https://anonymous.4open.science/r/graphSU-35B4}.

\end{abstract}

\section{Introduction} \label{sec:introduction}
Large language models (LLMs) can retain private, copyrighted, stale, or user-rescinded information after training. Exact unlearning would retrain the model on the corpus with the deletion set removed, but this is typically infeasible for modern LMs. Approximate LLM unlearning therefore aims to update a deployed model so that target knowledge is no longer recoverable, while its behaviour on retained data remains close to the original or retain-only model~\citep{maini2024tofu,shi2024muse,tian2024knowundo,li2024wmdp}.

As enterprises increasingly customize foundation models using proprietary organizational data, the ability to selectively remove previously learned information becomes critical. Data may later require removal due to privacy regulations, customer offboarding, intellectual-property concerns, policy revisions, or the discovery of sensitive content in training corpora. In such settings, machine unlearning provides a scalable mechanism for updating deployed models without incurring the computational and operational costs of full retraining.

Most current algorithms focus on the \emph{local objective}: gradient ascent, KL-regularised ascent, preference optimisation, negative preference optimisation, gradient rectification, energy-based unlearning, or representation control~\citep{jang2023knowledge,zhang2024npo,mekala2025altpo,wang2025gru,chaiyapattanaporn2026selu}. This line of work is valuable, but it leaves a complementary question under-addressed: \emph{where should unlearning be applied?} In selective unlearning, targeting only the canonical forget example is often insufficient, as the forgotten information may still be inferred from aliases, paraphrases, neighboring facts, or other training instances that describe the same entity from different perspectives. For example, forgetting \textit{``Acme Corp plans to acquire BetaAI for \$850 million''} might be insufficient if another sample states, \textit{``Acme's largest strategic investment this year is valued at approximately \$850 million''}. Conversely, applying unlearning too broadly can remove information that should be retained, leading to unnecessary degradation of model utility. This creates a scope-control problem that is distinct from the choice of unlearning objective.

In this paper, we formulate selective unlearning as a leakage-route control problem, where the goal is to prevent forgotten information from being reconstructed through paraphrases, aliases, and semantically related training instances while preserving unrelated knowledge. We propose \textbf{Graph}-guided \textbf{S}elective \textbf{U}nlearning (\method{}), a loss-agnostic controller that determines the fine-grained deletion scope before applying a local unlearning objective. \method{} constructs a weighted support graph over training instances using semantic similarity, symbolic entity/relation/tail overlap, and answer-side gradient alignment. It then propagates deletion pressure from forget seeds, identifies influential support neighbors, assigns soft forgetting strengths, and incorporates graph-selected neighbors through a staged unlearning curriculum. On GPT-2 Medium, \method{} achieves the lowest utility-feasible soft leakage across all deletion settings on both \textsc{TOFU} and \textsc{PISTOL}. Compared to a Seed-Only variant, it consistently reduces leakage while maintaining retain perplexity (PPL; the exponential of mean token-level negative log-likelihood, lower is better) below 10, demonstrating the importance of expanding beyond the initial forget set.

\section{Methodology}
\begin{figure*}[t]
  \centering
  \figmaybe[width=0.95\textwidth]{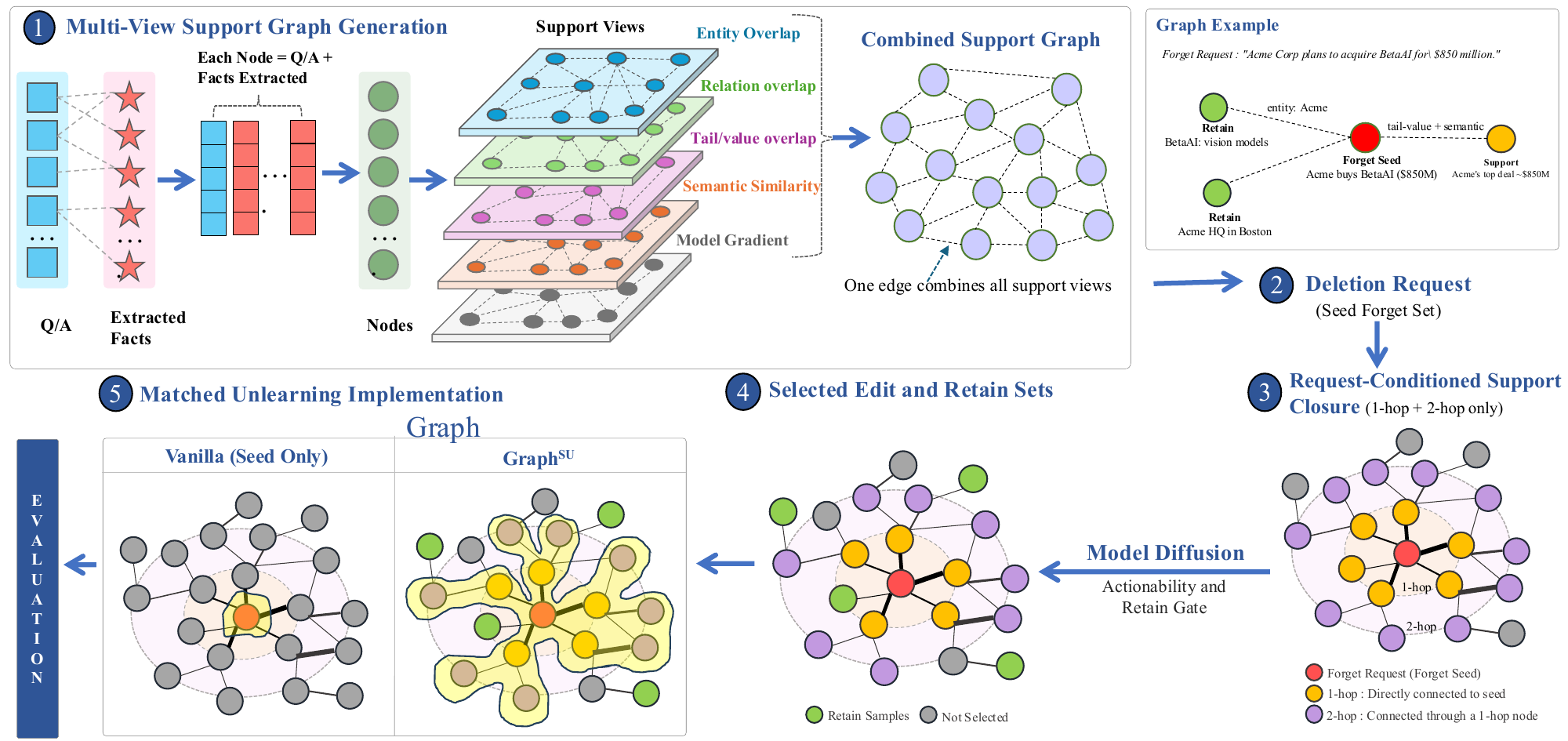}
  \caption{Overview of \method, which constructs a weighted multi-view support graph over QA samples, propagates deletion pressure from forget seeds through a one- and two-hop support closure, and applies graded unlearning objectives to the expanded edit set.}
  \Description{Workflow for graph-guided selective unlearning.}
  \label{fig:controller-flow}
\end{figure*}

Figure \ref{fig:controller-flow} illustrates proposed method.

\subsection{Problem Statement}



Let $f_{\theta_0}$ denote the original model trained on dataset  \( D=\{z_i\}_{i=1}^{N}\) where \(z_i=(q_i,a_i^{\mathrm{text}},\mathcal{F}_i)\).
Here, $q_i$ is the prompt, $a_i^{\mathrm{text}}$ is the answer surface form, and
\(\mathcal{F}_i=\{(h_{i,k},r_{i,k},t_{i,k})\}_{k=1}^{K_i}\)
is the set of factual units expressed in the answer. Each tuple
$(h_{i,k},r_{i,k},t_{i,k})$ consists of head entity, relation, and
tail entity or value. We obtain $\mathcal{F}_i$ by prompting GPT-4.1~\cite{openai2025gpt41} to extract factual predicate-argument tuples from the prompt--answer pair and normalizing the output into $(h,r,t)$ unit. The extracted units are used for target specification and support-graph construction, not as additional supervision for model training.

An unlearning request $\mathcal{R}$ identifies target fact units
$\mathcal{K}_{\mathcal{R}}$ to remove and a seed forget set
$D_f \subset D$ containing them; the initial retain pool is
$D_r = D \setminus D_f$. The goal of selective unlearning is to obtain an updated model
$f_{\theta_u}$ in which the knowledge specified by
$\mathcal{K}_{\mathcal{R}}$ is no longer recoverable, while the model's
utility on unrelated retain data is preserved.


\subsection{Recovery-Aware Deletion Scope}
\label{sec:deletion_scope}

As discussed in Section~\ref{sec:introduction},  the seed forget set $D_f$
may miss related training instances that still support recovery of the
target knowledge, leaving leakage routes outside $D_f$. We denote by
$\mathcal{M}_{\mathcal{R}}(f_\theta)$ a scalar recoverability measure
for $\mathcal{K}_{\mathcal{R}}$ under request-specific probes, capturing
the extent to which the model can still reproduce, paraphrase, or
semantically entail the target facts.

For any training example $z_j$, define its ideal support contribution as
\begin{equation}
\Delta_j^{\star}(\mathcal{R})
=
\mathcal{M}_{\mathcal{R}}(f_{\theta_0})
-
\mathcal{M}_{\mathcal{R}}(f_{\theta_0}^{(-j)}),
\end{equation}
where $f_{\theta_0}^{(-j)}$ denotes the model obtained by applying the
same supervised fine-tuning procedure after removing $z_j$ from the
training set.

We define the ideal support closure of $\mathcal{R}$ as
\begin{equation}
\mathcal{C}^{\star}_{\epsilon}(\mathcal{R})
=
D_f
\cup
\{z_j\notin D_f:\Delta_j^{\star}(\mathcal{R})\ge \epsilon\},
\end{equation}
where $\epsilon>0$ is a small operational threshold. This closure is the
ideal forgetting scope: it includes both the seed forget examples and
the non-seed examples whose removal would materially reduce recovery of
$\mathcal{K}_{\mathcal{R}}$. Constructing
$\mathcal{C}^{\star}_{\epsilon}(\mathcal{R})$ exactly is infeasible,
because it would require leave-one-out retraining or exact causal
attribution over the training corpus. \textsc{GraphSU} therefore
approximates this closure using a \textbf{request-conditioned support graph},
introduced next.

\subsection{Support Graph Construction.}
The seed forget set $D_f$ is not generally a closed evidence unit. Even after examples in $D_f$ are suppressed, the target knowledge may remain recoverable through
\textbf{paraphrastic}~\citep{maini2024tofu,reimers2019sentence,reisizadeh2025leakk},
\textbf{symbolic}~\citep{tian2024knowundo,uipe2025,wei2025forget,meng2022locating,zhong2023mquake,cohen2024evaluating},
and \textbf{model-internal} support routes~\citep{koh2017understanding,pruthi2020estimating,park2023trak}. We therefore construct a weighted support graph $G=(V,E,W)$, where each node corresponds to a training example and each edge weight estimates how strongly one example may support recovery of another. The graph is used as an affinity structure for support expansion, not as a calibrated estimate of causal influence.

\noindent\textbf{Multi-view support affinity.}
For a candidate edge $(i,j)$, \method{} fuses five support views three symbolic (entity, relation, tail), one semantic, and one gradient-based:
\begin{equation}
\label{eq:edge-weight}
\small
\begin{aligned}
\tilde{w}_{ij}&=\alpha s^{\mathrm{ent}}_{ij}+\beta s^{\mathrm{rel}}_{ij}+\gamma s^{\mathrm{tail}}_{ij}
+\delta s^{\mathrm{sem}}_{ij}+\lambda s^{\mathrm{grad}}_{ij},\\
w_{ij}&=\min\!\bigl(1,\max(0,\tilde{w}_{ij})\bigr),
\end{aligned}
\end{equation}
where each component score is normalized to $[0,1]$ and $(\alpha,\beta,\gamma,\delta,\lambda)$ are non-negative view weights (Table~\ref{tab:settings}). The five views capture complementary recovery routes: the three symbolic overlaps capture structured factual support through shared entities, relations, and tail values; semantic similarity captures paraphrastic support; and gradient alignment captures model-local support under the frozen post-SFT model.

\noindent\textbf{Semantic evidence.}
Let $e_i=\mathrm{Enc}([q_i;a_i])$
be the $\ell_2$-normalized sentence embedding of the prompt--answer pair. The semantic score is
\begin{equation}
\label{eq:sem-sim}
\small
s^{\mathrm{sem}}_{ij}
=
\left[\cos(e_i,e_j)\right]_+,
\qquad
[x]_+=\max(0,x).
\end{equation}
Negative semantic cosine indicates dissimilar rather than supporting content. Because \method{} models positive recovery support with a non-negative diffusion process, negative semantic affinities are clipped to zero rather than allowed to subtract from supportive edges. Modelling inhibitory relationships would require a signed-graph formulation and is left for future work.

\noindent\textbf{Symbolic evidence.}
Semantic similarity can miss structured dependencies such as aliases, inverse relations, and reused private values. We extract canonicalized entity, relation, and tail-value sets $\mathcal{E}_i$, $\mathcal{R}_i$, and $\mathcal{T}_i$ from $q_i$, $a_i$, and $\mathcal{F}_i$ respectively. Entity aliasing and relation canonicalization map equivalent mentions and inverse relations to shared identifiers. We then compute entity-, relation-, and tail-level symbolic similarities $s{_{ij}}$ between instances $i$ and $j$ using Jaccard similarity over the corresponding sets.

Entity overlap ($s^{\mathrm{ent}}$) captures subject- and alias-level support, relation overlap ($s^{\mathrm{rel}}$) captures factual-route support, and tail-value overlap ($s^{\mathrm{tail}}$) captures reused attributes or identifiers.

\noindent\textbf{Gradient-alignment evidence.}
The semantic and symbolic views are corpus-level approximations. To capture model-local support, we compute answer-side gradient summaries under the frozen post-SFT model. Let $\mathcal{A}_i$ denote the answer-token positions of $z_i$. We define
\begin{equation}
\label{eq:grad-summary}
g_i
=
\frac{1}{|\mathcal{A}_i|}
\sum_{t\in\mathcal{A}_i}
\nabla_{H^{(L)}_{it}}
\mathcal{L}_{\mathrm{CE}}(z_i;\theta_0),
\end{equation}
where $H^{(L)}_{it}$ is the final-layer answer representation after LayerNorm and before the LM head. The gradient-alignment score is
\begin{equation}
\label{eq:grad-sim}
s^{\mathrm{grad}}_{ij}
=
\left[\cos(g_i,g_j)\right]_+ .
\end{equation}
Positive gradient cosine indicates compatible, mutually reinforcing local update directions, whereas negative cosine indicates conflicting directions rather than recovery support. Since \method{} propagates non-negative support mass, negative gradient affinities are clipped to zero; representing inhibitory relations would instead require signed-graph propagation.

The graph captures recovery-aware support structure around the request. \method{} propagates deletion pressure from $D_f$ through $G$, assigning graded forgetting strengths to high-risk neighbors while leaving unrelated examples intact.



\subsection{Graph-Guided Deletion Scope Approximation}
Given an unlearning request $\mathcal{R}$ with the seed forget set $D_f$. We first estimate which non-seed examples are most
likely to belong to the recovery-aware deletion scope of the request.

Let $A$ denote the non-negative adjacency matrix of the sparsified support graph. We use the pairwise affinity $w_{ij}$ from Eq.~\ref{eq:edge-weight} directly, so
\begin{equation}
\small
A_{ij}=
\begin{cases}
w_{ij}, & (i,j)\in E \text{ after sparsification},\\
0, & \text{otherwise}.
\end{cases}
\end{equation}
Because the component similarities in Eq.~\ref{eq:edge-weight} are pairwise symmetric, $w_{ij}=w_{ji}$ before sparsification; the retained graph is treated as undirected for diffusion. Let $d_i=\sum_j A_{ij}$ be the weighted degree of node $i$. We define the row-normalized transition matrix
\begin{equation}
\small
P_{ij}
=
\begin{cases}
A_{ij}/d_i, & d_i>0,\\
\mathbb{I}[i=j], & d_i=0,
\end{cases}
\end{equation}
where isolated nodes are assigned self-loops. The transition matrix
specifies how support evidence moves across the graph: a node passes
more evidence to neighbors connected by stronger semantic, symbolic, or
gradient-aligned support edges. We initialize a request-specific seed distribution
\begin{equation}
\small
s_i=
\begin{cases}
1/|D_f|, & z_i\in D_f,\\
0, & z_i\notin D_f.
\end{cases}
\end{equation}
The support score vector $r$ is then computed by personalized graph
diffusion:
\begin{equation}
\label{eq:ppr}
\small
r^{(k+1)}
=
(1-\rho)P^\top r^{(k)}+\rho s,
\end{equation}
where $\rho\in(0,1)$ is the restart probability. The restart term anchors the diffusion
to the deletion request, while the transition term propagates support
evidence through the graph. At convergence, the solution satisfies
\begin{equation}
\small
\label{eq:ppr-fixed}
r
=
\rho
\left(I-(1-\rho)P^\top\right)^{-1}s ,
\end{equation}
or equivalently,
\begin{equation}
\small
\label{eq:path-expansion}
r
=
\rho
\sum_{\ell=0}^{\infty}
(1-\rho)^{\ell}
(P^\top)^{\ell}s .
\end{equation}
This expansion shows that $r_i$ aggregates support evidence over all
paths from the seed forget examples to $z_i$, with paths of length
$\ell$ discounted by $(1-\rho)^\ell$. Hence, $r_i$ is high when $z_i$ is
connected to $D_f$ through many strong short paths, or through multiple
consistent longer paths. The score is not interpreted as a probability
of membership in the deletion scope; it is a ranking signal for
request-specific support evidence.

The estimated support closure is obtained by selecting high-scoring
non-seed examples together with the original seeds. For a threshold
$\eta$ or an expansion budget $B$, we use
\begin{equation}
\label{eq:estimated-closure-threshold}
\small
\widehat{\mathcal{C}}_{\eta}(\mathcal{R})
=
D_f
\cup
\{z_i\in D_r: r_i\ge \eta\},
\end{equation}
or equivalently the top-$B$ non-seed examples ranked by $r_i$ when a
fixed expansion budget is required. This construction approximates the
ideal support closure
$\mathcal{C}^{\star}_{\epsilon}(\mathcal{R})$ defined in
Section~\ref{sec:deletion_scope}: examples with larger diffusion scores
are treated as more likely to materially support recovery of the target
knowledge.

\subsection{Selective Unlearning}
\label{sec:support-weighted-unlearning}
Let $D_{sup}(\mathcal{R})$ be the selected non-seed support set:
\begin{equation}
\label{eq:support-set}
\small
D_{\mathrm{sup}}(\mathcal{R})
=\widehat{\mathcal{C}}_{\eta}(\mathcal{R})\setminus D_f.
\end{equation}
Each example receives a forgetting weight
\begin{equation}
\label{eq:omega}
\small
\omega_i=
\begin{cases}
1, & z_i\in D_f,\\
\eta_{\mathcal{R}}
\left(
\dfrac{r_i}{\max_{z_j\in D_{\mathrm{sup}}} r_j}
\right)^{\gamma_\omega},
& z_i\in D_{\mathrm{sup}}(\mathcal{R}),\\
0, & \text{otherwise}.
\end{cases}
\end{equation}
Thus, seed examples receive full forgetting pressure, while non-seed support examples receive bounded pressure according to graph relevance. Let $m_i$ be a binary answer-token mask, with $m_{i,t}=1$ for target tokens to be forgotten and $m_{i,t}=0$ for answer tokens that should be retained. The complement $1-m_i$ therefore selects the safe answer tokens. The unlearning objective is
\begin{equation}
\label{eq:graphsu-risk}
\begin{aligned}
\mathcal{L}_{\mathrm{GraphSU}}
&= \sum_{i=1}^{N} \Big[
\omega_i\mathcal{L}_{\mathrm{forget}}(z_i,m_i) \\[-1mm]
&\qquad + \lambda_r\mathcal{L}_{\mathrm{retain}}(z_i,1-m_i)
\Big] \\[-1mm]
&\qquad + \lambda_{\mathrm{KL}}\mathcal{L}_{\mathrm{KL}}.
\end{aligned}
\end{equation}
The forgetting term is applied only to the target positions selected by $m_i$ and combines three components:
\begin{equation}
\label{eq:forget-loss}
\mathcal{L}_{\mathrm{forget}}
=
\lambda_{\mathrm{ent}}\mathcal{L}_{\mathrm{ent}}
+
\lambda_{\mathrm{UL}}\mathcal{L}_{\mathrm{UL}}
+
\lambda_{\mathrm{rep}}\mathcal{L}_{\mathrm{rep}}.
\end{equation}
Here, $\mathcal{L}_{\mathrm{ent}}$ is the entropy-maximization term that flattens the predictive distribution at target tokens, $\mathcal{L}_{\mathrm{UL}}$ is the unlikelihood term that suppresses probability assigned to the target tokens, and $\mathcal{L}_{\mathrm{rep}}$ is the representation-repulsion term used to move target-token representations away from their frozen post-SFT representations. The coefficients $\lambda_{\mathrm{ent}}$, $\lambda_{\mathrm{UL}}$, and $\lambda_{\mathrm{rep}}$ control the relative strengths of these terms. The retain loss $\mathcal{L}_{\mathrm{retain}}$ is cross-entropy on the complementary safe-token mask $1-m_i$, while $\mathcal{L}_{\mathrm{KL}}$ anchors non-target predictive behaviour to the frozen post-SFT model; $\lambda_r$ and $\lambda_{\mathrm{KL}}$ are their corresponding weights. Representation repulsion is enabled for complete and entity/instance deletion and disabled for partial deletion, where target and non-target facts may share the same answer context. In this way, \method{} separates the scope controller from the local unlearning loss: the graph determines which examples and tokens are edited, while the loss defines how the selected targets are suppressed.

\section{Experiments}
\subsection{Experimental Settings}
\noindent\textbf{Datasets and Models.}
We evaluate \method{} on two unlearning benchmarks: \textsc{TOFU}~\citep{maini2024tofu}, a synthetic author-profile QA benchmark with auditable forget/retain splits, and \textsc{PISTOL}~\citep{qiu2024pistol}, a structural-unlearning benchmark built around interconnected factual samples. We consider complete, entity/instance, and partial deletion settings. We use GPT-2 Medium for the main controlled comparison and \llamamodel{} (\llamacheckpoint{}) to assess scalability to larger backbones. For each model, all methods are initialized from the same supervised fine-tuned checkpoint and evaluated using identical forget seeds and prompts.




\noindent\textbf{Baselines.}
We compare \method{} with existing methods including GA~\cite{jang2023knowledge}, GA+KL~\cite{maini2024tofu}, DPO~\citep{rafailov2023dpo}, NPO-TR~\citep{zhang2024npo}, AltPO~\citep{mekala2025altpo}, GRU~\citep{wang2025gru}, and SELU~\citep{chaiyapattanaporn2026selu}. NPO-TR denotes an NPO-style baseline with an explicit retain-term regularizer.
We also compare \method{} with Seed-Only, a controlled ablation that shares the same unlearning objective as \method{} but removes graph-guided controller. Key implementation details are summarized in Table~\ref{tab:settings}.

\begin{table}[t]
\centering
\caption{Core implementation settings. Full hyperparameter sweeps are deferred to supplementary material.}
\label{tab:settings}
\scriptsize
\setlength{\tabcolsep}{4pt}
\begin{tabular}{lccc}
\toprule
Setting & Complete & Entity & Partial \\
\midrule
Steps & 1200 & 1200 & 720 \\
Graph activation $s_g$ & 120 & 120 & 220 \\
Neighbour cap $K_n$ & 72 & 72 & 40 \\
Neighbour $\phi^{(n)}_{max}$ & 0.35 & 0.32 & 0.15 \\
Entropy weight $\lambda_{\mathrm{ent}}$ & 1.35 & 1.40 & 1.15 \\
Unlikelihood weight $\lambda_{\mathrm{UL}}$ & 0.60 & 0.60 & 0.25 \\
Retain CE weight $\lambda_r$ & 0.22 & 0.20 & 0.24 \\
KL weight $\lambda_{\mathrm{KL}}$ & 0.01 & 0.01 & 0.012 \\
Repulsion $\lambda_{\mathrm{rep}}\mathcal{L}_{\mathrm{rep}}$ & on & on & off \\
Route prompts per seed & 4 & 5+aliases & 6 \\
\midrule
Edge weights $(\alpha,\beta,\gamma,\delta,\lambda)$ & \multicolumn{3}{c}{$(0.7,0.5,0.5,1.0,0.7)$} \\
\bottomrule
\end{tabular}
\end{table}

\subsection{Metrics and Route Probes}
The primary forget metric is soft leakage. For each forget seed $z_i\in D_f$, let $\{p_{i,k}\}_{k=1}^{K_i}$ denote its validated recovery-route prompts, where $k$ indexes an applicable route prompt and $K_i$ is the number of valid prompts for that seed after route generation and filtering (Appendix~\ref{app:route-prompts}). Let $y'_{i,k}$ be the model continuation generated from $p_{i,k}$, and let $\mathrm{target}_i$ denote the sensitive answer string or grounded target value whose recoverability is being tested for seed $i$. We define $s_{i,k}=\operatorname{sim}(y'_{i,k},\mathrm{target}_i)$ as the sentence-embedding cosine similarity between the generated continuation and that target. Soft leakage is
\begin{equation}
\softleak
=
\frac{1}{|D_f|}
\sum_{z_i\in D_f}
\frac{1}{K_i}
\sum_{k=1}^{K_i}
\mathbbm{1}\!\left[\,s_{i,k}\ge \tau_{\mathrm{soft}}\,\right],
\end{equation}
where $\tau_{\mathrm{soft}}=0.85$ is the similarity threshold used to count a route as leaking the target. For both datasets, we evaluate three deletion settings: complete deletion, where the target is the full answer; partial deletion, where the target is a set of sensitive spans; and entity deletion, where the target is the canonical entity profile together with its aliases. Retention is measured using retain perplexity (PPL), retain F1, ROUGE-L F1, and separate retain-neighbor, retain-far, and partial-retain-only panels. PPL is the exponential of the mean token-level negative log-likelihood, so lower values indicate better retained language-modelling performance. Appendix~\ref{app:utility} reports the additional utility panels jointly with soft leakage. Route families include direct, paraphrase, indirect, cloze, related-fact, \listsummary{}, \relinversion{}, and entity-alias prompts.

\subsection{Main Results}
\noindent\textbf{Utility-constrained unlearning.} Table~\ref{tab:main-gpt2} reports the main GPT-2 Medium results on
\textsc{TOFU} and \textsc{PISTOL}; results with \llamamodel{} are
provided in Appendix~\ref{app:llama}. Across both datasets and all deletion settings, \method{} achieves the lowest soft leakage among methods that keep retain PPL below 10. This distinction is important because effective unlearning requires balancing forgetting and retention. DPO, AltPO, and SELU preserve low retain PPL but leave high leakage, while GA, GA+KL, and GRU can reduce leakage only with substantial retention loss. For example, on \textsc{TOFU} complete deletion, GRU obtains lower leakage than \method{} (38.13\% vs.\ 46.83\%) but raises retain PPL to 79.11, making it unsuitable under the utility-feasible criterion.

\begin{table*}[t]
\centering
\scriptsize
\setlength{\tabcolsep}{4.0pt}
\renewcommand{\arraystretch}{1.08}
\begin{tabularx}{\textwidth}{ll Y Y  Y Y  Y Y}

\toprule
\textbf{Dataset} & \textbf{Method}
& \multicolumn{2}{c}{\textbf{Complete}}
& \multicolumn{2}{c}{\textbf{Entity/Instance}}
& \multicolumn{2}{c}{\textbf{Partial}} \\
\cmidrule(lr){3-4}
\cmidrule(lr){5-6}
\cmidrule(lr){7-8}
& & \textbf{Leak$\downarrow$} & \textbf{PPL$\downarrow$}
& \textbf{Leak$\downarrow$} & \textbf{PPL$\downarrow$}
& \textbf{Leak$\downarrow$} & \textbf{PPL$\downarrow$} \\
\midrule
\multirow{9}{*}{\textbf{\textsc{TOFU}}} & GA & 97.13 & 24.91 & 97.75 & 51.13 & 99.75 & 61.24 \\
& GA+KL & 97.13 & 24.91 & 97.75 & 51.13 & 88.25 & 42.44 \\
& DPO & 99.88 & \textbf{1.24} & 100.00 & \textbf{1.30} & 100.00 & 2.94 \\
& NPO-TR & 91.50 & 4.79 & 88.63 & 9.12 & 95.13 & 10.75 \\
& AltPO & 97.50 & 1.58 & 97.19 & \textbf{1.30} & 100.00 & 9.02 \\
& GRU & \textbf{$38.13^{\dagger}$} & 79.11 & 65.31 & 59.09 & 86.25 & 24.42 \\
& SELU & 92.88 & 1.29 & 97.88 & 1.32 & 98.75 & \textbf{1.27} \\
& Seed-Only & 93.25 & 2.61 & 96.25 & 4.40 & 100.00 & 7.90 \\
\rowcolor{gray!18}
& \textbf{\method{} (Proposed)} & \underline{46.83} & 3.27 & \textbf{\underline{54.60}} & 4.94 & \textbf{\underline{81.67}} & 1.91 \\
\midrule
\multirow{9}{*}{\textbf{\textsc{PISTOL}}} & GA & 100.00 & \textbf{1.02} & 100.00 & 1.05 & 100.00 & 1.20 \\
& GA+KL & 100.00 & \textbf{1.02} & 99.39 & 1.06 & 99.34 & 1.20 \\
& DPO & 99.83 & \textbf{1.02} & 100.00 & \textbf{1.02} & 100.00 & 1.28 \\
& NPO-TR & 98.92 & \textbf{1.02} & 94.50 & 1.10 & 99.17 & 1.39 \\
& AltPO & 95.33 & 1.15 & 99.50 & 1.03 & 63.67 & 3.13 \\
& GRU & 99.87 & 1.04 & 98.70 & 1.06 & 98.76 & 1.37 \\
& SELU & 96.66 & \textbf{1.02} & 100.00 & 1.03 & 97.78 & \textbf{1.01} \\
& Seed-Only & 56.33 & 1.09 & 26.33 & 1.10 & 41.39 & 1.22 \\
\rowcolor{gray!18}
& \textbf{\method{} (Proposed)} & \textbf{\underline{6.83}} & 1.05 & \textbf{\underline{4.67}} & 1.10 & \textbf{\underline{29.17}} & 1.19 \\
\bottomrule
\end{tabularx}
\caption{Main forget--retain comparison on GPT-2 Medium across \textsc{TOFU} and \textsc{PISTOL}. Leak denotes soft leakage (\%); PPL denotes retain perplexity; lower is better for both. Bold marks the best value in each column, regardless of method. Underlining marks the lowest-leakage result satisfying the operational retention constraint PPL $\leq 10$. $\dagger$ marks the lower-leakage \textsc{TOFU} complete-deletion result that violates this constraint.}
\label{tab:main-gpt2}
\end{table*}

\noindent\textbf{Impact of graph-guided support expansion.} 
Since Seed-Only uses the same unlearning objective but omits graph-neighbor expansion, the comparison isolates the effect of support-route control in \method{}. On \textsc{TOFU}, \method{} reduces aggregate soft leakage by 46.42 points for complete deletion, 41.65 points for entity/instance deletion, and 18.33 points for partial deletion. On \textsc{PISTOL}, the corresponding reductions are 49.50, 21.66, and 12.22 points. Figure~\ref{fig:route_heatmap} further shows that the improvement is consistent across recovery-route families: \method{} improves over Seed-Only in all 21 task--route cells, with average route-level reductions of 45.9, 41.7, and 22.6 points for complete, entity/instance, and partial deletion, respectively. These results show that effective unlearning requires deleting support routes, not just seed examples. Appendix~\ref{app:utility} complements PPL with retain F1, ROUGE-L F1, far-retain F1, neighbour-retain F1, and safe-span F1 for partial deletion. On \textsc{TOFU}, \method{} raises retain F1 from 39.07 to 42.41 for complete deletion, from 37.16 to 59.11 for entity/instance deletion, and from 3.20 to 55.14 for partial deletion; under partial deletion, safe-span F1 rises from 1.26 to 61.31. These answer-level results show that the controlled Seed-Only gains reflect an improved forgetting--utility balance rather than leakage reduction in isolation. 

\noindent\textbf{Challenges of partial deletion.}
Complete and entity/instance deletion benefit most from graph expansion because target information can be supported through full answers, aliases, and related facts. Partial deletion is more constrained: the model must suppress sensitive spans while preserving the remaining answer content. Although \method{} improves the partial-deletion frontier, achieving 81.67\% leakage with retain PPL 1.91 on \textsc{TOFU}, substantial residual leakage remains. These results support graph-guided support expansion and route-level control, while stronger claims about certified erasure require further ablation.

\subsection{Recovery-route robustness}
\begin{figure}[t]
\centering
\includegraphics[scale=0.85]{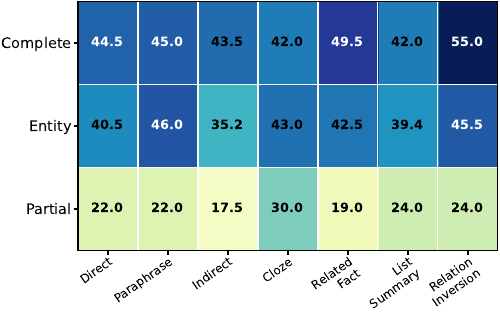}
\caption{Route-level soft-leakage reduction of \textsc{GraphSU} relative to Seed-Only on GPT-2 Medium (\textsc{TOFU}). Values denote percentage-point reductions in soft leakage; higher is better.}
\label{fig:route_heatmap}
\end{figure}

Figure~\ref{fig:route_heatmap} reports the route-level soft-leakage reduction of \textsc{GraphSU} relative to Seed-Only across seven recovery-route families and three deletion settings. \textsc{GraphSU} improves over Seed-Only in all 21 task--route combinations, reducing leakage by an average of 36.8 percentage points. Gains are largest for complete deletion, with an average reduction of 45.9 points and a maximum of 55.0 points under relation-inversion prompts. Entity deletion shows a similar pattern, averaging 41.7 points with improvements of at least 35.2 points across all routes. Partial deletion yields smaller but consistent gains (22.6 points on average), reflecting the need to preserve non-sensitive answer content while suppressing sensitive spans. Since both methods use the same local objective and route prompts, these results provide controlled evidence that graph-guided support expansion improves robustness to diverse recovery routes.

\section{Related Work}
Early LLM unlearning studies adapted gradient ascent or retain-regularised objectives to suppress forget examples~\citep{jang2023knowledge,maini2024tofu}. More recent methods address known failure modes of direct ascent: NPO slows catastrophic collapse through a preference-inspired loss~\citep{zhang2024npo}; AltPO combines negative and positive preference signals to avoid nonsensical forget responses~\citep{mekala2025altpo}; GRU rectifies gradients to mitigate the unlearning--retention trade-off~\citep{wang2025gru}; and SELU uses an energy-based LoRA formulation with straight-through estimators~\citep{chaiyapattanaporn2026selu}. These algorithms mainly improve the update rule. \method{} is complementary: it determines the support routes, masks, and strengths to which an update rule is applied.


\section{Conclusions}
We presented \method{}, a graph-guided controller that treats selective unlearning as a support-route control problem: instead of editing only the explicitly identified forget seeds, it builds a weighted multi-view support graph, propagates request-conditioned relevance from the seeds, and applies graded forgetting strengths to the high-risk neighbours. \method{} can be paired with existing unlearning objectives without modifying them. Across \textsc{TOFU} and \textsc{PISTOL} with GPT-2 Medium and \llamamodel{}, \method{} achieves the lowest utility-feasible soft leakage in all deletion settings, reducing leakage by up to 49.5 percentage points over a matched Seed-Only ablation while maintaining low retain perplexity. These results suggest that the deletion scope, not only the unlearning loss, is a key determinant of reliable knowledge removal.

\section*{Limitations}
Unlike complete or entity deletion, partial deletion requires removing sensitive spans while preserving the surrounding answer content, making the task inherently more constrained. A well-posed partial-deletion instance requires an answer that contains at least two clearly separable facts, one sensitive and targeted for removal, and one safe and retained so that the sensitive span can be suppressed without disturbing the remainder. This is a property of the span-level deletion setting itself rather than of any particular method, and the residual leakage observed in this setting largely reflects this structural difficulty.

More broadly, soft leakage measures behavioural recoverability under a fixed family of recovery-route probes rather than certified representational erasure, so our results are best interpreted as evidence of substantially reduced reachability of the target knowledge. \method{} additionally performs offline fact extraction and support-graph construction; this cost is incurred once per corpus and amortised across deletion requests (Appendix~\ref{app:ops}).

\section*{Ethical Considerations}
Selective unlearning can support privacy, copyright compliance, removal of stale information, and safer model maintenance. It can also create a false sense of compliance if evaluation is weak, or be misused to remove safety-relevant knowledge. We therefore present \method{} as an approximate unlearning technique that should be paired with governance, audit trails, and independent verification.

\bibliography{references}

\clearpage
\appendix
\section{Experiment Results on \llamamodel{}}
\label{app:llama}

\noindent\textbf{Larger-backbone evaluation.}
We evaluate \method{} with \llamamodel{} on \textsc{TOFU} to examine whether graph-guided support expansion transfers beyond GPT-2 Medium. All methods start from the same supervised fine-tuned checkpoint and use the same deletion manifests, evaluation prompts, decoding settings, and checkpoint-selection criterion. Lower values are better for both soft leakage and retain perplexity.

\noindent\textbf{Complete deletion.}
\begin{table}[ht]
\centering
\scriptsize
\setlength{\tabcolsep}{6pt}
\begin{tabular}{lrr}
\toprule
Method & Leak$\downarrow$ & PPL$\downarrow$ \\
\midrule
Seed-Only & 93.67 & \textbf{1.05} \\
GA        & 99.70 & 1.08 \\
GA+KL     & 94.33 & 1.09 \\
DPO       & 94.33 & 1.30 \\
NPO-TR    & 91.66 & 1.06 \\
AltPO     & 90.00 & 1.08 \\
GRU       & 93.33 & 1.08 \\
SELU      & 93.66 & \textbf{1.05} \\
\rowcolor{gray!18}
\textbf{\method{}} & \textbf{13.33} & 1.96 \\
\bottomrule
\end{tabular}
\end{table}

\noindent\textbf{Entity/instance deletion.}
\begin{table}[ht]
\centering
\scriptsize
\setlength{\tabcolsep}{6pt}
\begin{tabular}{lrr}
\toprule
Method & Leak$\downarrow$ & PPL$\downarrow$ \\
\midrule
Seed-Only & 25.11 & 1.04 \\
GA        & 36.00 & 1.20 \\
GA+KL     & 38.66 & 1.05 \\
DPO       & 92.22 & 1.70 \\
NPO-TR    & 33.33 & 1.09 \\
AltPO     & 34.67 & 1.15 \\
GRU       & 86.33 & 1.03 \\
SELU      & 35.66 & \textbf{1.02} \\
\rowcolor{gray!18}
\textbf{\method{}} & \textbf{15.53} & 1.35 \\
\bottomrule
\end{tabular}
\end{table}

\noindent\textbf{Partial deletion.}
\begin{table}[ht]
\centering
\scriptsize
\setlength{\tabcolsep}{6pt}
\begin{tabular}{lrr}
\toprule
Method & Leak$\downarrow$ & PPL$\downarrow$ \\
\midrule
Seed-Only & 94.33 & 1.56 \\
GA        & 94.36 & 1.05 \\
GA+KL     & 89.00 & 1.16 \\
DPO       & 96.36 & \textbf{1.03} \\
NPO-TR    & 89.66 & 1.07 \\
AltPO     & 88.67 & 1.09 \\
GRU       & 93.67 & 1.06 \\
SELU      & 94.66 & 1.04 \\
\rowcolor{gray!18}
\textbf{\method{}} & \textbf{87.01} & 1.07 \\
\bottomrule
\end{tabular}
\end{table}

The larger-backbone results preserve the central pattern observed in the controlled GPT-2 Medium experiments: expanding the edit scope beyond the explicit forget seeds is most beneficial when the target remains reachable through related examples or entity-centred support routes. Partial deletion remains more difficult because the method must suppress only a sensitive span while preserving the remainder of the answer.

\noindent\textbf{Perplexity consistency check.} Retain PPL is computed by teacher-forced scoring on the same frozen retain set for a given comparison. Prompt and padding positions are masked, answer-token negative log-likelihood (NLL) is summed over the evaluation set and divided by the number of scored answer tokens, and the aggregate mean NLL is exponentiated once. Consequently, PPL must be at least one. The DPO partial-deletion value previously shown as 0.03 was mean NLL rather than PPL; the corrected value is $\exp(0.03)\approx 1.03$. This correction does not change the leakage ranking because DPO soft leakage is 96.36\%.

\noindent\textbf{Applicability beyond QA-formatted data and causal language models.}
GPT-2 Medium and \llamamodel{} are autoregressive causal language models (CLMs), not QA-specific architectures; the QA pairs used here are the benchmarks' data representation. More generally, a \method{} node can represent a document, passage, or conversation, with deletion masks identifying sensitive spans, while the same semantic, symbolic, and gradient support views define request-conditioned support routes for non-QA corpora.

\section{Pareto Trade-Off Plots}
\label{app:pareto}
The plots below show the forget--retain trade-off for the three evaluated model--dataset settings. Lower-left is better; the shaded region marks the utility-feasible region, and the dashed line denotes retain PPL $=10$ where shown. To preserve legibility in the two-column EMNLP format, each model--dataset plot is rendered as a separate full-width vector figure.

The PPL $\leq 10$ threshold is used only as a fixed operational guardrail to prevent severe retention collapse from being interpreted as successful unlearning; it is not presented as a universal cutoff. We therefore also inspect the unfiltered Pareto frontier, retaining every reported operating point regardless of PPL.

\begin{table}[ht]
\centering
\caption{Unfiltered Pareto-optimal points on GPT-2 Medium \textsc{TOFU}. Each pair is (soft leakage \%, retain PPL); lower is better on both axes.}
\label{tab:pareto-unfiltered}
\scriptsize
\setlength{\tabcolsep}{3pt}
\begin{tabularx}{\linewidth}{@{}>{\raggedright\arraybackslash}p{0.25\linewidth}X@{}}
\toprule
\textbf{Setting} & \textbf{Pareto-optimal operating points} \\
\midrule
Complete & DPO (99.88, 1.24); SELU (92.88, 1.29); \method{} (46.83, 3.27); GRU (38.13, 79.11) \\
Entity/instance & AltPO (97.19, 1.30); Seed-Only (96.25, 4.40); \method{} (54.60, 4.94) \\
Partial & SELU (98.75, 1.27); \method{} (81.67, 1.91) \\
\bottomrule
\end{tabularx}
\end{table}

\method{} is Pareto-optimal in all three deletion settings. It has the lowest leakage among the reported methods under every PPL cap from 5 through 50. With no retention cap, GRU attains lower complete-deletion leakage (38.13\%) but at retain PPL 79.11. NPO-TR is only slightly above the original guardrail for partial deletion (PPL 10.75), yet its soft leakage remains 95.13\% compared with 81.67\% for \method{}. The guardrail should therefore be read as one practical operating region rather than as a filter that defines the underlying trade-off.

\begin{figure*}[!t]
\centering
\figmaybe[width=0.32\textwidth]{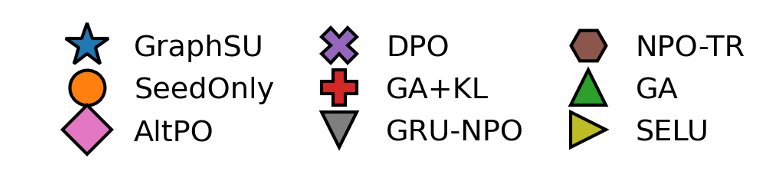}\\[3pt]
\figmaybe[width=0.96\textwidth]{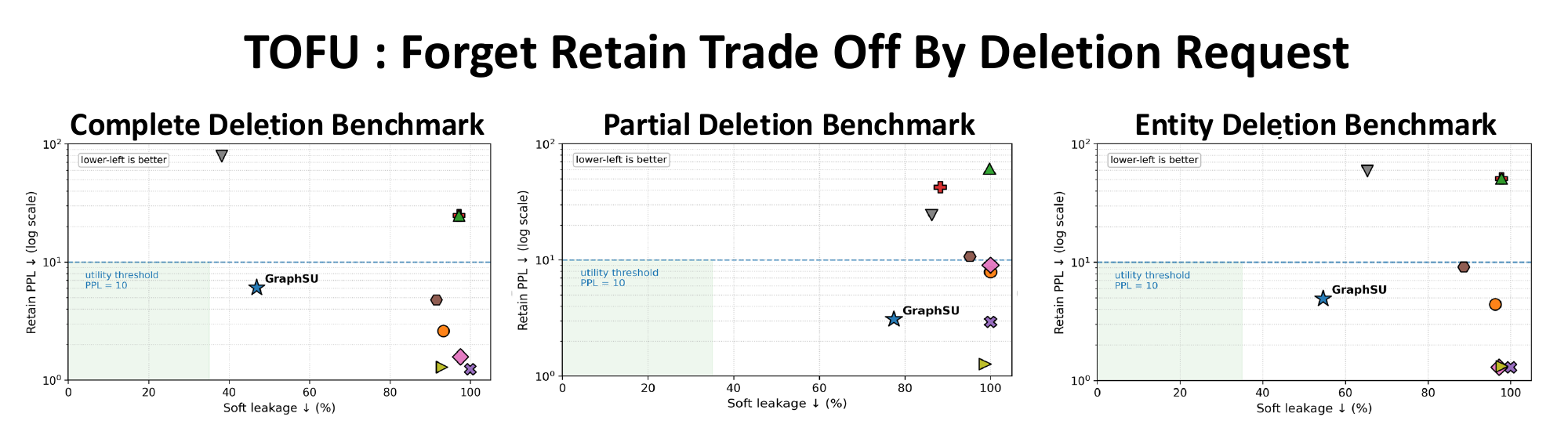}
\caption{Forget--retain trade-off for GPT-2 Medium on \textsc{TOFU}. The complete, partial, and entity/instance deletion panels are shown at full width so axis labels, method labels, and the utility-feasible region remain readable.}
\label{fig:pareto-tofu-gpt2}
\end{figure*}

\begin{figure*}[!t]
\centering
\figmaybe[width=0.32\textwidth]{figures/pareto_legend.png}\\[3pt]
\figmaybe[width=0.96\textwidth]{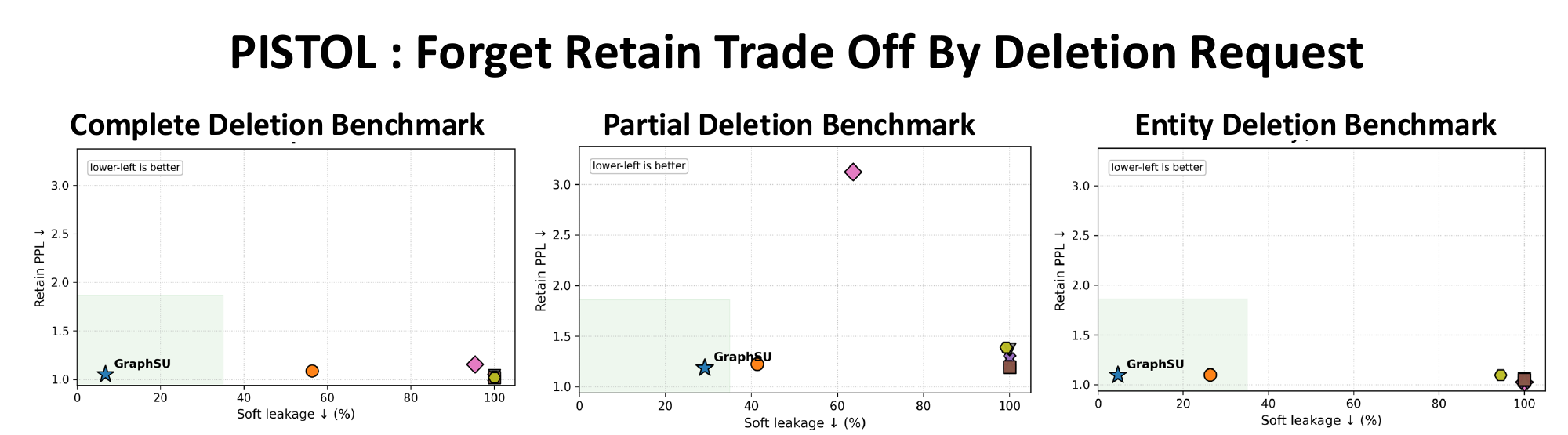}
\caption{Forget--retain trade-off for GPT-2 Medium on \textsc{PISTOL}, shown as a full-width vector panel for improved readability.}
\label{fig:pareto-pistol-gpt2}
\end{figure*}

\begin{figure*}[!t]
\centering
\figmaybe[width=0.32\textwidth]{figures/pareto_legend.png}\\[3pt]
\figmaybe[width=0.96\textwidth]{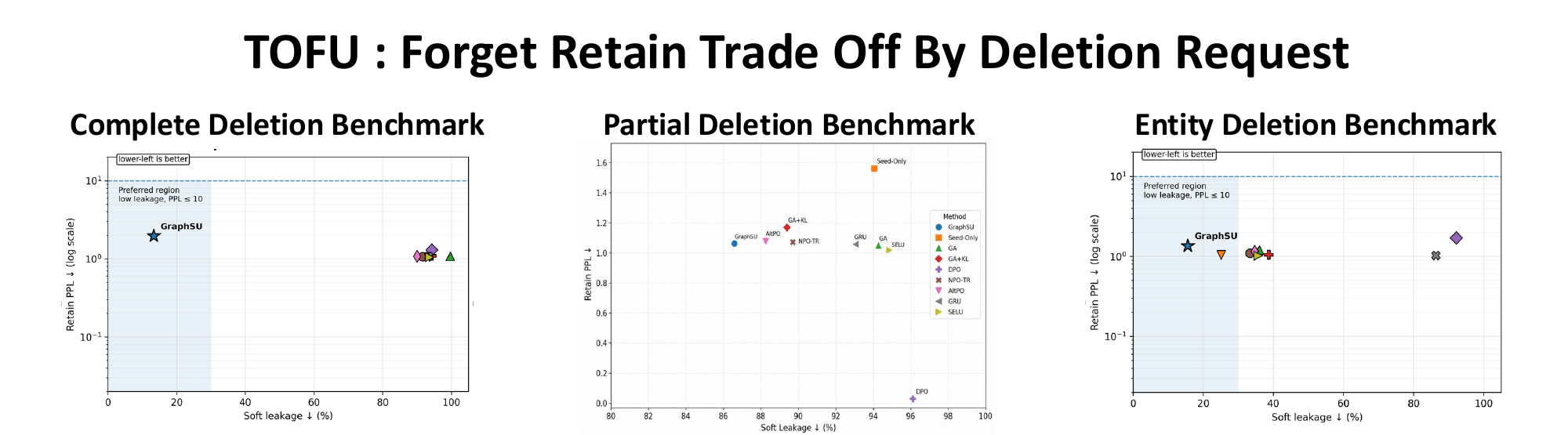}
\caption{Forget--retain trade-off for \llamamodel{} on \textsc{TOFU}, shown as a full-width vector panel for improved readability.}
\label{fig:pareto-tofu-llama}
\end{figure*}

\section{Fact Extraction and GPT-4.1 Prompting}
\label{app:fact-extraction}
\method{} uses GPT-4.1 only for offline fact grounding. For each QA pair, GPT-4.1 extracts schema-constrained head--relation--tail facts. The outputs are then validated, canonicalised, deduplicated, and type-checked using deterministic post-processing. Extraction is performed once per corpus and cached; it is not repeated for each deletion request. The recovery-route prompts in Appendix~\ref{app:route-prompts} are generated locally and do not use an external LLM.

\begin{table}[ht]
\centering
\caption{Fact-extraction settings.}
\label{tab:extraction-settings}
\scriptsize
\begin{tabularx}{\linewidth}{@{}>{\raggedright\arraybackslash}p{0.42\linewidth}X@{}}
\toprule
\textbf{Setting} & \textbf{Value} \\
\midrule
Extraction model & GPT-4.1 \\
Output format & JSON object (\texttt{json\_object} mode) \\
Nominal workload & Approximately one initial request per QA sample \\
Maximum retry attempts & 3 \\
Request timeout & 60 seconds \\
Maximum labelled few-shot examples & 20 \\
\bottomrule
\end{tabularx}
\end{table}

\noindent\textbf{System instruction.}
\begin{lstlisting}[style=prompt]
You are an information extraction assistant that prepares data for a knowledge graph. Be CONSERVATIVE and ACCURATE: never invent books, awards, dates, or relationships that are not clearly stated in the question or answer. For each question/answer pair, extract ATOMIC factual triples suitable for graph edges. Each triple is an object with:
- "head": the entity the fact is about
- "head_type": one of [PERSON, ORG, WORK, EVENT, LOCATION, AWARD, GENRE, CHARACTER, TOPIC, OTHER]
- "relation": a short, descriptive edge label
- "tail": the target node, value, or linked entity
- "tail_type": one of [PERSON, ORG, WORK, EVENT, LOCATION, DATE, NUMBER, OCCUPATION, AWARD, GENRE, CHARACTER, TOPIC, NATIONALITY, LANGUAGE, OTHER]
BOOKS/WORKS: create a separate triple per titled work; use PERSON--author_of-->WORK; quoted strings in a works context are strong WORK candidates; add WORK--genre-->GENRE and WORK--publication_year-->DATE when stated. LANGUAGE/NATIONALITY: tag tail_type accordingly.
Use both question and answer; split comma/"and" lists into separate triples; do not invent facts. Return ONLY: {"triples": [...]}
\end{lstlisting}

\noindent\textbf{Per-sample user prompt and worked example.}
\begin{lstlisting}[style=prompt]
Study the supplied canonical examples, then extract graph triples for:
Question: Where does Avery Collins work?
Answer: Avery Collins works at Northbridge University and writes in Spanish.

-> {"triples": [
  {"head":"Avery Collins","head_type":"PERSON",
   "relation":"works_at","tail":"Northbridge University","tail_type":"ORG"},
  {"head":"Avery Collins","head_type":"PERSON",
   "relation":"writes_in","tail":"Spanish","tail_type":"LANGUAGE"}
]}
\end{lstlisting}

\section{Local Fact Extraction and Error Propagation}
\label{app:local-extraction}
Hosted fact extraction may be undesirable for proprietary corpora. We therefore rebuilt the symbolic graph with the local Qwen2.5-3B-Instruct model and repeated a matched \textsc{TOFU} complete-deletion experiment. Only extractor-derived facts and symbolic edges changed; GPT-2 Medium, all 184 forget targets, training code, seed 42, unlearning recipe, evaluation manifest, decoding, and detector thresholds were held fixed.

Qwen processed 4,005 QA examples and returned valid triples for 3,942 (98.43\%) without external API calls. In a 200-example structured audit, exact-triple F1 was 15.57\% for Qwen and 9.88\% for GPT-4.1; GPT-4.1 retained higher exact-tail accuracy. Because semantically equivalent facts can be decomposed into different triples, this structured audit is diagnostic rather than a complete semantic gold standard.

The two rebuilt graphs remained close despite extractor differences: edge-weight Spearman correlation was $\rho=0.925$, degree correlation was $\rho=0.979$, diffusion-score correlation was $\rho=0.996$, and the selected support sets had Jaccard similarity 0.895 (68 of 72 selected supports were identical). To test error propagation, we additionally corrupted graph-active heads, tails, and triples. At 40\% random-active corruption, 69 of 72 support nodes were retained; under 40\% graph-critical corruption, 66 of 72 were retained. Diffusion-score correlations remained at least 0.9974. These perturbations indicate that symbolic noise is attenuated by the semantic and gradient views, weighted fusion, and diffusion rather than being automatically propagated to the full support closure.

\begin{table}[ht]
\centering
\caption{Matched step-480 \textsc{TOFU} complete-deletion operating point using local Qwen2.5-3B-Instruct fact extraction. The Original-SFT column is the common pre-unlearning checkpoint.}
\label{tab:local-qwen}
\scriptsize
\setlength{\tabcolsep}{2.5pt}
\begin{tabular}{lrr}
\toprule
\textbf{Metric} & \textbf{Original-SFT} & \textbf{Local-Qwen \method{}} \\
\midrule
Soft leakage $\downarrow$ & 91.25\% & \textbf{49.83\%} \\
Hard leakage $\downarrow$ & 37.67\% & \textbf{1.33\%} \\
Forget-answer F1 $\downarrow$ & 84.37\% & \textbf{33.86\%} \\
Retain F1 $\uparrow$ & 72.55\% & 30.27\% \\
Retain semantic cosine $\uparrow$ & 90.95\% & 79.53\% \\
Retain PPL $\downarrow$ & 4.30 & 9.95 \\
\bottomrule
\end{tabular}
\end{table}

At this matched operating point, local extraction gives soft leakage 49.83\%, approximately 3.0 percentage points above the re-verified GPT-4.1-based \method{} result (46.83\%), while avoiding external transmission of the corpus. Leakage decreases by 36--44 percentage points across every tested route family and hard leakage is 0--3\%. The trade-off is not cost-free: lexical retain utility falls, and neighbour-retain PPL reaches 11.54 even though the three-panel aggregate retain PPL is 9.95. We therefore view local extraction as a privacy-preserving deployment option with a measurable forgetting--utility trade-off, not as an accuracy-equivalent replacement for GPT-4.1.

\section{Adversarial Route-Prompt Generation}
\label{app:route-prompts}
For each evaluated sample $z_i=(q_i,a_i^{\mathrm{text}},\mathcal{F}_i)$, we select a grounded target fact $k_i=(h_i,r_i,t_i)$, where $h_i$ is the head entity, $r_i$ is the relation, and $t_i$ is the target value whose recoverability is assessed. Recovery-route prompts are generated offline using deterministic, relation-conditioned templates; no external LLM is used in this stage.

The generator may additionally use a validated alias set $A(h_i)$, an inverse-relation mapping $I(r_i)$, and related facts associated with the same entity or a graph-connected sample. Candidate prompts are normalised, deduplicated, checked for missing template slots, accidental target disclosure, and relation compatibility. The resulting manifests are frozen before evaluation and reused unchanged for \method{} and all baselines. The task-specific monitoring panels use the prompt counts in Table~\ref{tab:settings}; not every route is applicable to every target, and validated aliases may add entity-specific variants.

\subsection{Detailed Worked Example}
Consider the illustrative QA sample:
\begin{quote}
\textbf{Question:} Where does Avery Collins work?\\
\textbf{Answer:} Avery Collins works at Northbridge University.
\end{quote}
The fact-extraction stage produces the grounded target fact with head \emph{Avery Collins}, relation \texttt{works\_at}, and tail \emph{Northbridge University}.
For this example, the generator is also provided with the validated aliases
\begin{equation}
A(h_i)=\{\text{A. Collins},\text{Avery C.}\},
\end{equation}
and, when available, a related fact such as
\begin{equation}
(\text{Avery Collins},\texttt{researches},\text{machine ethics}).
\end{equation}
All prompts below assess recovery of the same target value, Northbridge University; only the recovery route changes.

After template expansion, prompts pass through the following deterministic validation procedure:
\begin{enumerate}
\item fill route-specific slots using the grounded target fact, validated aliases, and available related facts;
\item remove prompts with missing or invalid slots;
\item remove prompts that reveal the target value, except when the route definition requires it, as in relation inversion;
\item remove exact and normalised duplicates;
\item retain only prompts compatible with the target relation;
\item order or sample valid variants using the fixed experimental seed; and
\item save the resulting manifest for reuse across all evaluated methods.
\end{enumerate}
Thus, differences between \method{} and the baselines cannot be attributed to different evaluation questions: each method uses the same frozen target facts, route prompts, decoding settings, and soft-leakage threshold.

\begin{table*}[t]
\centering
\caption{Recovery-route families and representative generation rules. The displayed prompts are illustrative; the implementation uses relation-conditioned template maps so that each prompt remains compatible with its grounded target relation.}
\label{tab:route-families}
\scriptsize
\begin{tabularx}{\textwidth}{@{}>{\raggedright\arraybackslash}p{0.13\textwidth}>{\raggedright\arraybackslash}p{0.35\textwidth}X@{}}
\toprule
\textbf{Route family} & \textbf{Generation rule} & \textbf{Representative prompt} \\
\midrule
Direct & Preserve the original question, optionally adding a neutral instruction wrapper. & ``Please answer: $\langle q_i\rangle$'' \\
Paraphrase & Apply relation-specific lexical and syntactic transformations while preserving $h_i$, $r_i$, and target $t_i$. & ``At which institution is $\langle h_i\rangle$ employed?'' \\
Indirect & Ask for the same relation using contextual or descriptive wording rather than the original surface form. & ``Which organisation is associated with $\langle h_i\rangle$'s employment?'' \\
Cloze & Convert the target fact into a declarative statement and replace the target value with a blank. & ``$\langle h_i\rangle$ works at \underline{\hspace{1.5em}}.'' \\
Related-fact & Introduce another validated fact associated with $h_i$ or a selected graph neighbour, and then query the target relation. & ``The researcher who studies $\langle\text{related topic}\rangle$ is affiliated with which institution?'' \\
Entity alias & Replace the canonical entity mention with a validated alias from $A(h_i)$. & ``Where does $\langle\mathrm{alias}(h_i)\rangle$ work?'' \\
List-summary & Request several profile attributes while ensuring that the target relation is included. & ``List the main profile details for $\langle h_i\rangle$, including institutional affiliation.'' \\
Relation inversion & Reverse the grounded relation using $I(r_i)$ when the inverse is meaningful and sufficiently unambiguous. & ``Which author is affiliated with $\langle t_i\rangle$?'' \\
\bottomrule
\end{tabularx}
\end{table*}

\begin{table*}[t]
\centering
\caption{Detailed route generation for one grounded target fact.}
\label{tab:route-example}
\scriptsize
\begin{tabularx}{\textwidth}{@{}>{\raggedright\arraybackslash}p{0.12\textwidth}>{\raggedright\arraybackslash}p{0.42\textwidth}X@{}}
\toprule
\textbf{Route} & \textbf{Generated prompt} & \textbf{Purpose} \\
\midrule
Direct & Where does Avery Collins work? & Tests recovery through the original question. \\
Paraphrase & At which institution is Avery Collins employed? & Changes lexical and syntactic form while preserving the target relation. \\
Indirect & Which university is associated with Avery Collins's employment? & Avoids the original ``where does ... work'' formulation. \\
Cloze & Avery Collins works at \underline{\hspace{1.5em}}. & Tests whether the target completes a factual statement. \\
Related-fact & The researcher known for work on machine ethics is affiliated with which institution? & Uses a related fact to reach the target entity; emitted only when that fact identifies the intended entity without ambiguity. \\
Entity alias & Where does A. Collins work? & Tests recovery after replacing the canonical entity name with a validated alias. \\
List-summary & List Avery Collins's main profile details, including institutional affiliation. & Tests whether the target appears within a broader profile response. \\
Relation inversion & Which author is affiliated with Northbridge University? & Reverses the relation direction; emitted only when the inverse mapping is defined and sufficiently unambiguous. \\
\bottomrule
\end{tabularx}
\end{table*}

\section{Taxonomy-Blind and Held-Out Leakage Probes}
\label{app:heldout-probes}
The templated route families in Appendix~\ref{app:route-prompts} are useful for controlled diagnosis, but robustness measured only on that taxonomy can overstate generalisation. We therefore perform an additional taxonomy-blind evaluation on all 184 \textsc{TOFU} complete-deletion targets. A local Qwen2.5-3B-Instruct model generates free-form probes using only the original question and answer. It receives no graph structure, route labels, templates, hyperparameters, or outputs from the unlearned models. The resulting probes have zero exact matches with the original template bank. Checkpoints and detector thresholds are frozen before this evaluation.

We evaluate 736 free-form probes (four per target) and a stronger attack panel of 342 probes across 171 targets, where each probe is first verified to recover the target answer from Original-SFT. Seed-Only and \method{} use identical prompts and decoding. Leakage is measured with (i) a fixed-reference representation derived from Original-SFT, (ii) external MPNet semantic similarity, (iii) DeBERTa natural-language inference, (iv) direct target-phrase recovery, and (v) a conservative combined detector that fires when any constituent detector succeeds.

\begin{table*}[t]
\centering
\caption{Taxonomy-blind leakage on unseen free-form probes for \textsc{TOFU} complete deletion. Reduction is Seed-Only minus \method{} in percentage points. Confidence intervals are paired target-level 95\% bootstrap intervals; $p$ is the two-sided McNemar test.}
\label{tab:heldout-taxonomy-blind}
\scriptsize
\setlength{\tabcolsep}{5pt}
\begin{tabular}{llrrrrl}
\toprule
\textbf{Panel} & \textbf{Detector} & \textbf{Seed-Only} & \textbf{\method{}} & \textbf{Reduction} & \textbf{95\% CI} & \textbf{McNemar $p$} \\
\midrule
All targets & Fixed-reference & 74.18\% & \textbf{61.68\%} & 12.50 pp & [8.02, 17.26] & $1.03\times10^{-8}$ \\
All targets & Combined & 75.68\% & \textbf{62.09\%} & 13.59 pp & [8.97, 18.34] & $5.98\times10^{-10}$ \\
Confirmed attacks & Fixed-reference & 78.95\% & \textbf{65.20\%} & 13.74 pp & [7.02, 20.76] & $4.92\times10^{-5}$ \\
Confirmed attacks & Combined & 79.82\% & \textbf{65.50\%} & 14.33 pp & [7.89, 21.05] & $2.23\times10^{-5}$ \\
\bottomrule
\end{tabular}
\end{table*}

The direction is also consistent for direct phrase recovery, MPNet, and NLI considered separately. For continuity with the main evaluation, the original soft-leakage detector is retained as an auxiliary measure: on the full taxonomy-blind panel it favours \method{} (93.48\% versus 96.20\% for Seed-Only; $p=0.0187$), while it saturates on the smaller confirmed-attack panel. These results extend the claim from robustness within the predefined route family to improved robustness against unseen, independently generated recovery attempts; they do not constitute certified erasure.

\section{Qualitative Deletion Examples}
\label{app:qualitative}
The examples below use only the concepts defined in the main paper: the deletion setting, forget seed, grounded target fact, edit scope, support routes, and retention constraint. They illustrate how requests are represented and evaluated rather than introducing a separate response-transformation framework.

\subsection{Complete deletion}
\noindent\textbf{Forget seed.} \emph{Question:} Where does Avery Collins work? \emph{Answer:} Avery Collins works at Northbridge University. The grounded target has head \emph{Avery Collins}, relation \texttt{works\_at}, and tail \emph{Northbridge University}.

\noindent\textbf{Edit scope and interpretation.} All answer tokens in the forget seed receive full forgetting pressure, while graph-selected support examples receive bounded pressure according to their request-conditioned support scores. A successful update prevents recovery of Northbridge University through applicable direct, paraphrase, cloze, related-fact, list-summary, and relation-inversion prompts while preserving utility on unrelated retain examples.

\subsection{Entity/instance deletion}
\noindent\textbf{Deletion request.} Remove the profile associated with Hsiao Yun-Hwa. One seed asks: \emph{What is the full name of the author born in Taipei, Taiwan on 05/11/1991 who writes in the leadership genre?} The seed answer is: \emph{The author's full name is Hsiao Yun-Hwa.}

\noindent\textbf{Edit scope and interpretation.} Canonical-name, alias, and entity-profile examples form the seed forget set, and additional high-risk support examples are selected through the request-conditioned graph closure. A successful update prevents recovery of the target profile through canonical-name, alias, indirect, related-fact, and list-summary prompts while preserving factual responses for unrelated entities. The target is suppressed without introducing a manually selected generic identity.

\subsection{Partial deletion}
\noindent\textbf{Forget seed.} \emph{Question:} Where does Avery Collins work, and in which language do they write? \emph{Answer:} Avery Collins works at Northbridge University and writes in Spanish. The sensitive target is the \texttt{works\_at} value \emph{Northbridge University}; the safe fact is that Avery Collins writes in Spanish.

\noindent\textbf{Edit scope and interpretation.} Forgetting is localised to the target span, while the safe answer span remains supervised; representation repulsion is disabled for this setting. A successful update prevents recovery of the institutional affiliation across applicable routes while retaining the model's ability to answer that Avery Collins writes in Spanish.

These examples clarify the three deletion granularities. Complete deletion suppresses the full answer associated with a seed; entity/instance deletion expands the request across an entity-centred profile and its aliases; and partial deletion suppresses only the grounded sensitive span while preserving the safe answer remainder.

\subsection{Diagnosing Residual Leakage in the Support Graph}
\label{app:leak-diagnostics}
The support graph localises potential support at the \emph{training-example level}; it does not claim to identify a single neuron or parameter in which a fact is stored. When a target remains recoverable after unlearning, the request-conditioned graph provides a diagnostic view through each node's diffusion score $r_i$, hop distance from the forget seed, and the semantic, symbolic, and gradient views contributing to its incident edges. Residual leakage can arise when relevant evidence lies just below the selection threshold or top-$B$ expansion budget, when support is distributed across several individually weak paths, or when partial deletion intentionally limits forgetting pressure because sensitive and safe facts share the same answer context. Thus, difficult-to-remove information may correspond to redundant or diffuse support routes rather than a single missed example. This interpretation is diagnostic rather than a calibrated causal localisation claim.

\section{Operational Cost, Memory, and Scalability}
\label{app:ops}
Unless otherwise stated, the measurements below use GPT-2 Medium on \textsc{TOFU}. Corpus-level facts and graph artefacts are cached and reused across deletion requests.

\subsection{Graph-construction scaling benchmark}
We benchmark nested \textsc{TOFU} subsets using GPT-2 Medium. GPT-4.1 fact extractions were already cached and are therefore excluded from the build-time measurements. Embedding and symbolic processing ran on the CPU, while gradient-alignment features used the GPU. The graph degree is capped at 30.

\begin{table*}[!t]
\centering
\caption{Graph construction, memory, cache size, and request-conditioned selection cost as the number of stored QA samples increases. Warm selection reuses in-memory cached artefacts; cold selection includes loading cached graph artefacts before support selection.}
\label{tab:graph-scaling}
\scriptsize
\setlength{\tabcolsep}{4.5pt}
\begin{tabular}{r r r r r r r r r}
\toprule
\textbf{Samples} & \textbf{Stored edges} & \textbf{Avg. degree} & \makecell{\textbf{Build}\\\textbf{time (s)}} & \makecell{\textbf{CPU RAM}\\\textbf{(GB)}} & \makecell{\textbf{GPU mem.}\\\textbf{(GB)}} & \makecell{\textbf{Cache}\\\textbf{(MB)}} & \makecell{\textbf{Warm sel.}\\\textbf{(s)}} & \makecell{\textbf{Cold sel.}\\\textbf{(s)}} \\
\midrule
1K & 11,279 & 22.56 & 106.01 & 1.41 & 2.36 & 2.87 & 0.68 & 1.27 \\
2K & 22,795 & 22.80 & 212.41 & 1.45 & 2.47 & 4.93 & 1.74 & 3.02 \\
4K & 45,524 & 22.76 & 452.36 & 1.40 & 2.36 & 7.82 & 4.01 & 6.53 \\
\bottomrule
\end{tabular}
\end{table*}

At 4,000 samples, graph construction takes 452.36 seconds (about 7.5 minutes), cached artefacts occupy 7.82 MB, and warm support selection takes 4.01 seconds. Because the stored degree is capped at $K=30$, the number of stored edges is $O(NK)$ and therefore linear in $N$ for fixed $K$; graph storage and each sparse diffusion iteration scale with the stored edge set. The current exact cosine candidate-retrieval step is approximately quadratic in the number of samples and may become the dominant bottleneck at much larger corpus sizes. Approximate nearest-neighbour retrieval for the semantic view and symbolic inverted indices are natural scalable replacements that preserve the sparse downstream graph.

\subsection{Request-time and audit costs}
The isolated warm/cold timings in Table~\ref{tab:graph-scaling} measure graph loading and support selection on cached subset artefacts. The full request-conditioned pipeline additionally includes request grounding, one- and two-hop retrieval, support scoring, and retain-safety filtering.

\begin{table}[ht]
\centering
\caption{Stage-level operational costs from the full deletion workflow.}
\label{tab:ops-stages}
\scriptsize
\begin{tabularx}{\linewidth}{@{}>{\raggedright\arraybackslash}p{0.31\linewidth}X@{}}
\toprule
\textbf{Stage} & \textbf{Observed cost or workload} \\
\midrule
Request-conditioned support selection & 11.4--33.4 seconds per deletion request (mean 25.9 seconds), including request grounding, one- and two-hop support retrieval, support scoring, and retain-safety filtering. No external calls are used. \\
Model update and orchestration & Approximately 5.7--6.9 minutes per request, measured as total wall time minus logged evaluation time. The residual includes setup and checkpointing and is not a pure training-time measurement. \\
Full recovery-route audit & 63.0--83.1 minutes and 4,464--5,376 local model generations. This is an offline validation cost rather than request-time latency and uses no external LLM calls. \\
\bottomrule
\end{tabularx}
\end{table}

With cached corpus artefacts, the recurring graph-specific overhead is small relative to model updating and the offline recovery-route audit. Fact extraction and graph construction are one-time corpus-level preprocessing costs and are amortised across deletion requests.

\section{Additional Utility Evaluation Beyond PPL}
\label{app:utility}
Perplexity is useful for detecting broad language-model degradation, but it does not fully capture whether retained factual behaviour is preserved. We therefore interpret soft leakage jointly with retain F1, ROUGE-L F1, far-retain F1, neighbour-retain F1, and safe-span F1 for partial deletion. The PPL $\leq 10$ criterion used in the main paper is an operational guardrail for these experiments rather than a universal unlearning threshold.

\begin{table}[H]
\centering
\caption{Primary \textsc{TOFU} forgetting and utility metrics for the controlled Seed-Only comparison.}
\label{tab:additional-utility}
\scriptsize
\setlength{\tabcolsep}{3.1pt}
\begin{tabular}{llrrrr}
\toprule
\textbf{Setting} & \textbf{Method} & \makecell{\textbf{Soft}\\\textbf{leak}$\downarrow$} & \makecell{\textbf{Retain}\\\textbf{PPL}$\downarrow$} & \makecell{\textbf{Retain}\\\textbf{F1}$\uparrow$} & \makecell{\textbf{ROUGE-L}\\\textbf{F1}$\uparrow$} \\
\midrule
Complete & Seed-Only & 93.25 & \textbf{2.61} & 39.07 & 35.03 \\
Complete & \method{} & \textbf{46.83} & 3.27 & \textbf{42.41} & \textbf{47.36} \\
Entity & Seed-Only & 96.25 & \textbf{4.40} & 37.16 & 33.38 \\
Entity & \method{} & \textbf{54.60} & 4.94 & \textbf{59.11} & \textbf{56.82} \\
Partial & Seed-Only & 100.00 & 7.90 & 3.20 & 3.01 \\
Partial & \method{} & \textbf{81.67} & \textbf{1.91} & \textbf{55.14} & \textbf{51.07} \\
\bottomrule
\end{tabular}
\end{table}

\begin{table}[H]
\centering
\caption{Locality-sensitive retain metrics corresponding to Table~\ref{tab:additional-utility}. Safe-span F1 applies only to partial deletion.}
\label{tab:additional-local-utility}
\scriptsize
\setlength{\tabcolsep}{3.2pt}
\begin{tabular}{llrrr}
\toprule
\textbf{Setting} & \textbf{Method} & \makecell{\textbf{Far-retain}\\\textbf{F1}$\uparrow$} & \makecell{\textbf{Neighbour}\\\textbf{F1}$\uparrow$} & \makecell{\textbf{Safe-span}\\\textbf{F1}$\uparrow$} \\
\midrule
Complete & Seed-Only & 41.05 & 31.79 & -- \\
Complete & \method{} & \textbf{51.06} & \textbf{32.77} & -- \\
Entity & Seed-Only & 43.42 & 34.69 & -- \\
Entity & \method{} & \textbf{57.13} & \textbf{46.17} & -- \\
Partial & Seed-Only & 1.14 & 0.83 & 1.26 \\
Partial & \method{} & \textbf{54.58} & \textbf{54.54} & \textbf{61.31} \\
\bottomrule
\end{tabular}
\end{table}

The controlled comparison shows a stronger forgetting--utility balance for \method{} in all three settings. Under complete deletion, GRU reaches lower soft leakage (38.13\%) than \method{} but raises retain PPL to 79.11, illustrating why leakage should not be interpreted without retention. Under partial deletion, NPO-TR has a slightly higher safe-span F1 (64.97) than \method{} (61.31), but its retain PPL is 10.75 and soft leakage remains 95.13\%. Neighbour-retain F1 is preserved or improved by \method{} relative to Seed-Only: 32.77 vs. 31.79 for complete deletion, 46.17 vs. 34.69 for entity/instance deletion, and 54.54 vs. 0.83 for partial deletion.

\balance
\section{Ablation Coverage and Hyperparameter Scope}
\label{app:ablation-scope}
The controlled Seed-Only comparison is an ablation of the graph-guided scope controller as a whole: it uses the same local unlearning objective but omits graph-neighbour expansion. The local-extractor and corruption experiments in Appendix~\ref{app:local-extraction} further perturb the symbolic component while leaving the semantic and gradient views available. Together, these experiments establish that graph-guided support expansion matters and that the fused graph is robust to substantial symbolic noise.

The present experimental archive does not contain a complete leave-one-view-out sweep for all five affinity terms or independent performance sweeps over the fusion weights, restart probability $\rho$, hop limit, and expansion budget $B$. We therefore do not attribute the observed gains to any single graph view and do not claim that the reported hyperparameters are optimal. The evaluated configuration uses all five views, the fusion weights reported in Table~\ref{tab:settings}, one- and two-hop support retrieval, and the task-specific neighbour caps in Table~\ref{tab:settings}. A full factorial sensitivity analysis would require additional training runs and remains an important direction for establishing which views are necessary under different data regimes.

\section{Evaluation Consistency and Statistical Reliability}
\label{app:statistical-reliability}
All reported retain PPL values use the same teacher-forced protocol within a comparison: prompt and padding positions are masked, answer-token NLL is accumulated over the frozen retain set, the total is divided by the number of scored answer tokens, and the resulting mean NLL is exponentiated once. This definition implies PPL $\geq 1$ and motivated the correction in Appendix~\ref{app:llama}. The local-extractor experiment additionally reports the Original-SFT retain PPL (4.30) at its matched operating point.

The unlearning runs reported in the paper use one training seed. We therefore do not describe evaluation resampling as across-seed variance. For the independently generated taxonomy-blind probes, Table~\ref{tab:heldout-taxonomy-blind} reports paired target-level bootstrap confidence intervals and two-sided McNemar tests because the same probes are evaluated under both checkpoints. The available aggregate benchmark outputs do not support a defensible reconstruction of across-seed variance for all baselines; accordingly, no such variance is fabricated. Training-seed sensitivity remains a limitation of the current empirical study. Future multi-seed comparisons should use paired target-level intervals for leakage, paired example-level intervals for retain F1/ROUGE-L, and example-resampled aggregate NLL before exponentiation for PPL, with multiplicity correction when several baselines are tested.
\end{document}